\documentclass[conference]{IEEEtran}
\IEEEsettopmargin{t}{72pt}
\IEEEoverridecommandlockouts
\usepackage{cite}
\usepackage{amsmath,amssymb,amsfonts}
\usepackage{algorithmic}
\usepackage{graphicx}
\usepackage{textcomp}
\usepackage{xcolor}
\usepackage{subcaption}
\usepackage{booktabs}
\def\BibTeX{{\rm B\kern-.05em{\sc i\kern-.025em b}\kern-.08em
    T\kern-.1667em\lower.7ex\hbox{E}\kern-.125emX}}

\usepackage{color,soul} 

\definecolor{red}{rgb}{1.00,0.00,0.00}
\definecolor{blue}{rgb}{0.00,0.00,1.00}
\definecolor{green}{rgb}{0.30, 0.50,0.00}

\usepackage{hyperref}

\begin{document}

\title{\LARGE \bf
Single-Shot 6DoF Pose and 3D Size Estimation for \\
Robotic Strawberry Harvesting
}

\author{
Lun Li and Hamidreza Kasaei
\thanks{
Lun Li and Hamidreza Kasaei both are with Department of Artificial Intelligence, Bernoulli Institute, Faculty of Science and Enginerring, University of Groningen, The Netherlands
}
\thanks{
Emails: \{alan.li, hamidreza.kasaei\}@rug.nl
}
}

\maketitle

\begin{abstract}

In this study, we introduce a deep-learning approach for determining both the 6DoF pose and 3D size of strawberries, aiming to significantly augment robotic harvesting efficiency. Our model was trained on a synthetic strawberry dataset, which is automatically generated within the Ignition Gazebo simulator, with a specific focus on the inherent symmetry exhibited by strawberries. By leveraging domain randomization techniques, the model demonstrated exceptional performance, achieving an 84.77\% average precision (AP) of 3D Intersection over Union (IoU) scores on the simulated dataset. Empirical evaluations, conducted by testing our model on real-world datasets, underscored the model's viability for real-world strawberry harvesting scenarios, even though its training was based on synthetic data. The model also exhibited robust occlusion handling abilities, maintaining accurate detection capabilities even when strawberries were obscured by other strawberries or foliage. Additionally, the model showcased remarkably swift inference speeds, reaching up to 60 frames per second (FPS). 

\end{abstract}

\section{Introduction}

Strawberries, recognized as one of the most consumed fruits worldwide, hold substantial commercial implications. Despite their popularity, strawberries are labor-intensive, with labor costs making up 40\% of the total production expense, primarily attributed to the harvesting process \cite{hernandez2023current}. Seasonal labor shortages and escalating labor costs pose imminent challenges for the strawberry industry. Fortunately, the inherent characteristics of strawberries – their compact plant morphology and striking red fruit appearance – render them promising candidates for automation in production. The prospect of developing robots specialized in autonomous strawberry harvesting has garnered significant attention in the agricultural robotics community. Yet, the absence of a commercially viable, large-scale strawberry-picking robot for field or greenhouse use underscores the unresolved issues \cite{defterli2016review}. Among them, delicately handling the soft and fragile strawberries without causing any damage remains a paramount concern. 

\begin{figure}[htpb]
  \centering
  \includegraphics[width=0.485\textwidth]{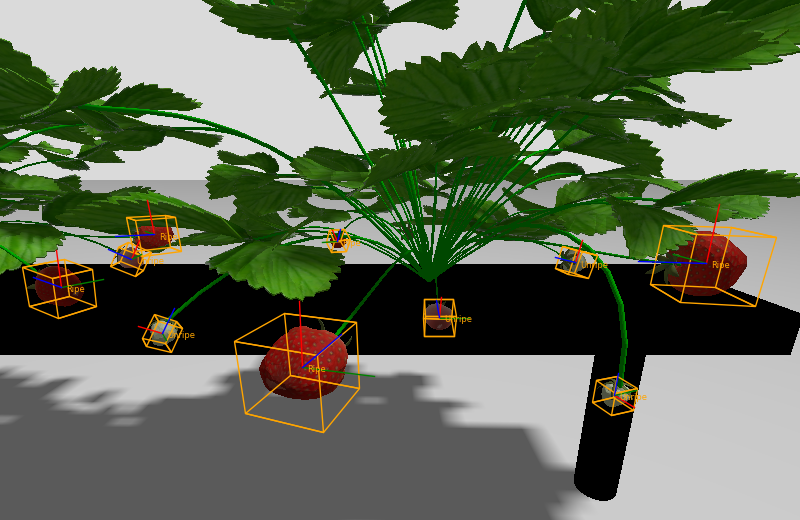}

  \vspace{1mm}
  \begin{minipage}{0.24\textwidth}
    \centering
    \includegraphics[width=\textwidth]{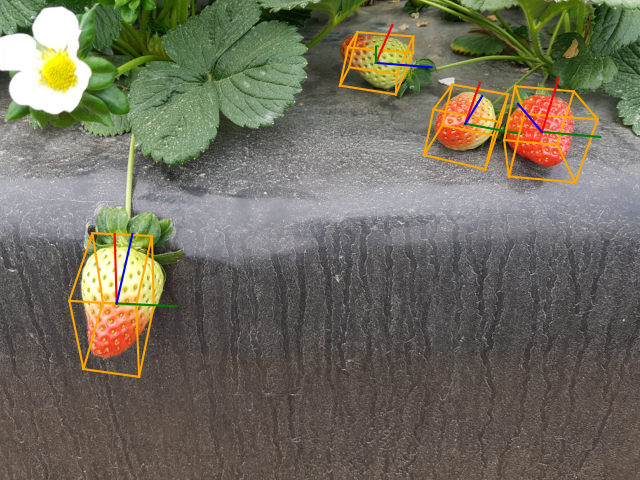}

  \end{minipage}
  \begin{minipage}{0.24\textwidth}
    \centering
    \includegraphics[width=\textwidth]{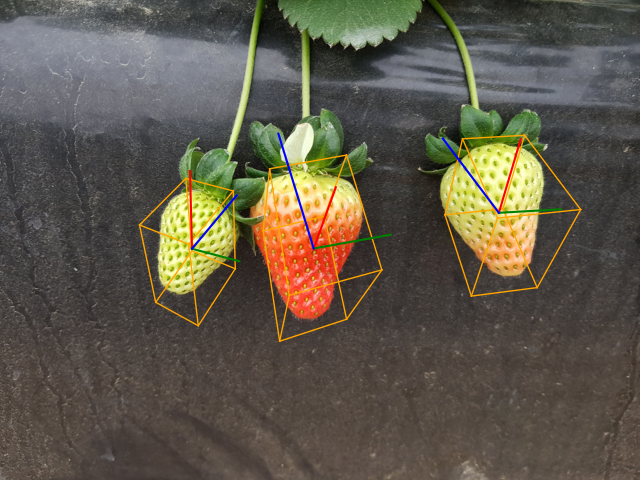}
  \end{minipage}
  
  \caption{We proposed a 6DoF pose and 3D size estimator capable of estimating the size and pose of all strawberry instances in a single view simultaneously. Notably, it could identify strawberries that were previously unseen, without reliance on any specific strawberry CAD models. 
  To train our model, we created a synthetic dataset, named Straw6D, tailored for 6DoF pose and 3D size estimation of strawberries (\textit{top-row}). Furthermore, our model demonstrated sim-to-real transferability to real-world images (\textit{lower-row}).}
  \label{fig1}
\end{figure}

In addressing the noted difficulty, this study aims to enhance robot perception to yield more precise and comprehensive strawberry localization information. Most extant robotic vision methods for strawberry picking employ a 2D to 3D transformation scheme\cite{montoya2022vision}. These methods commonly leverage either traditional image processing algorithms or learning-based techniques to first identify the 2D coordinates of strawberries within an image, and then correlate this data with depth information from specialized sensors to produce approximate 3D coordinates. However, regardless of how adeptly the fault-tolerant end-effector is crafted in subsequent stages, the overall performance of such robotic systems remains suboptimal due to the lack of complete 3D pose information. More explicitly, a full understanding of the 6 degrees of freedom (6DoF) pose of the target strawberry and its 3D size, is critical for a robotic arm to separate the target strawberry both safely and efficiently from dense clusters.

Although 6DoF pose estimation has long been a focal point in computer vision research, its implementation in real-world strawberry harvesting tasks remains challenging \cite{fan2022deep}. First and foremost, the majority of current methodologies concentrate on instance-level 6DoF pose estimation. Throughout both the training and inference phases, these approaches require a specific CAD model, disregarding intra-class variations. Such approaches are ill-suited for strawberries identification as the natural variability in strawberry shapes and sizes. 
Additionally, the strawberry picking environment presents significant occlusion challenges. Target strawberries are often obscured by other strawberries, plant stems, or even the robot itself, creating difficulties in estimating the strawberries' 6DoF poses from the sensory perspective of the robot.
Furthermore, state-of-the-art algorithms\cite{bruns2023rgb}, while exhibiting superior performance, are characterized by their high computational demands. The extensive consumption of computational resources and extended runtimes are not conducive for applications demanding efficiency and real-time responses, especially on resource-constrained edge computing platforms such as strawberry-harvesting robots.

Besides, while datasets hold equal importance to algorithms and computational capabilities, the manual annotation of a dataset on 6DoF pose emerges as an extremely laborious endeavor \cite{hinterstoisser2013model}. Unlike 2D visual tasks such as semantic segmentation, which can still be well executed manually, there is no naturally visible precise perspective view for 6DoF pose annotation. Consequently, untrained individuals often resort to extensive conjectures and auxiliary calibration tools. This not only amplifies the workload required for annotation but also precipitates a considerable decline in annotation quality. To overcome the challenges mentioned above, we present a suite of novel contributions:

\begin{itemize}
\item A computationally efficient categorical 6DoF pose estimation network designed exclusively for deriving both the 6DoF pose and 3D size of strawberries from RGB images, with due consideration to handle challenge on occlusion.
\item A simulation-driven automatic synthetic dataset generation and annotation pipeline for generating 6DoF pose and 3D sieze of strawberry samples with a holistic range of randomized lighting conditions, poses, sizes, maturities and occlusion configurations. The dataset is available at \href{https://huggingface.co/datasets/lishoulun/Straw6D}{\url{https://huggingface.co/datasets/lishoulun/Straw6D}}

\item Validating the sim-to-real transferability of the proposed approach on a real strawberry dataset~\cite{PEREZBORRERO2020105736}.

\end{itemize}

\section{Related Work}

Six degree-of-freedom (6DoF) pose estimation stands as a pivotal challenge within the realms of computer vision and robotics, captivating significant interest. This task is committed to ascertaining the 6DoF pose of an object within three-dimensional (3D) space through RGB or RGB-D images, including both 3D translations and 3D orientations. The evolution of this field has seen a pronounced shift towards the utilization of deep neural networks for addressing this problem, culminating in the emergence of two predominant methodologies: direct 6D pose regression from the input image, and two-phase strategies that initially forecast 2D-3D correspondences to subsequently deduce pose-related parameters.

Among the notable advancements, PoseCNN\cite{xiang2017posecnn} exemplifies an innovative convolutional neural network (CNN) architected for an end-to-end 6D pose estimation, ingeniously integrating a 3D coordinate regression network with a preemptive version of the RANSAC algorithm to secure robust estimations. Similarly, YOLO6D \cite{tekin2018real} adapts the famed object detection framework YOLO to provide a real-time, single-shot solution, utilizing a CNN to regress 2D projected coordinates before recovering the complete 6DoF pose with the assistance of the PnP algorithm and a CAD model. Meanwhile, DenseFusion \cite{wang2019densefusion}, distinguishes itself with a comprehensive approach that processes RGB-D images by concurrently addressing RGB and depth information, significantly enhancing the richness of its pose estimation insights.

In parallel, the introduction of streamlined neural network architectures, such as EfficientNet\cite{tan2019efficientnet}, mark a clear shift in research focus, emphasizing a balanced pursuit of model precision alongside computational efficiency. This balance is essential for real-world applications where computational resources are often constrained.

When the exploration of 6DoF pose estimation extends into the domain of agricultural robotics, a unique set of challenges are accentuated by the intricate nature of organic entities and the variability of outdoor settings. Complications such as occlusions, variable lighting, and the deformation of objects are commonplace. A critical consideration is the adaptability to natural variations, as exemplified by fruits, which seldom conform to a single CAD model. This necessitates a progression from instance-level to category-level 6DoF pose estimation, mirroring advancements in this area, such as NOCS\cite{wang2019normalized} and subsequent related research.

Kim et al. \cite{kim2022tomato} enriches this domine by introducing Deep-ToMaToS, a network imbued with a transformation loss, specifically devised for the 6DoF pose estimation of tomatoes and their attached side-stems with maturity classification. This methodology has been validated in both virtual and imitational smart farm environments, emphasizing its practical applicability. Nevertheless, this approach presumes uniformity in the size of the tomatoes, thus maintaining an instance-level perspective. In addition, Costanzo et al. \cite{costanzo2023enhanced} present a refined 6DoF pose estimation method designed for robotic fruit picking, accommodating apples of various dimensions and improving the efficiency of the harvesting process. Nonetheless, they carried out category-level through post-processing optimization, rather than through a direct end-to-end network modification.

In the context of strawberry harvesting, prevailing methods predominantly employ 2D techniques, such as the mask-rcnn, utilized in Tafuro et al. \cite{tafuro2022strawberry}. The advancement toward 6
DoF pose estimation for strawberries is still in its early phases, with Wagner et al. \cite{wagner2021efficient} leveraging the VGG network to predict the fruit's orientation successfully. Our research propels this further, innovating a categorical 6DoF pose estimation strategy for strawberries.

\section{Proposed Approaches}




\begin{figure*}[htpb]
\centering
\includegraphics[width=0.90\textwidth]{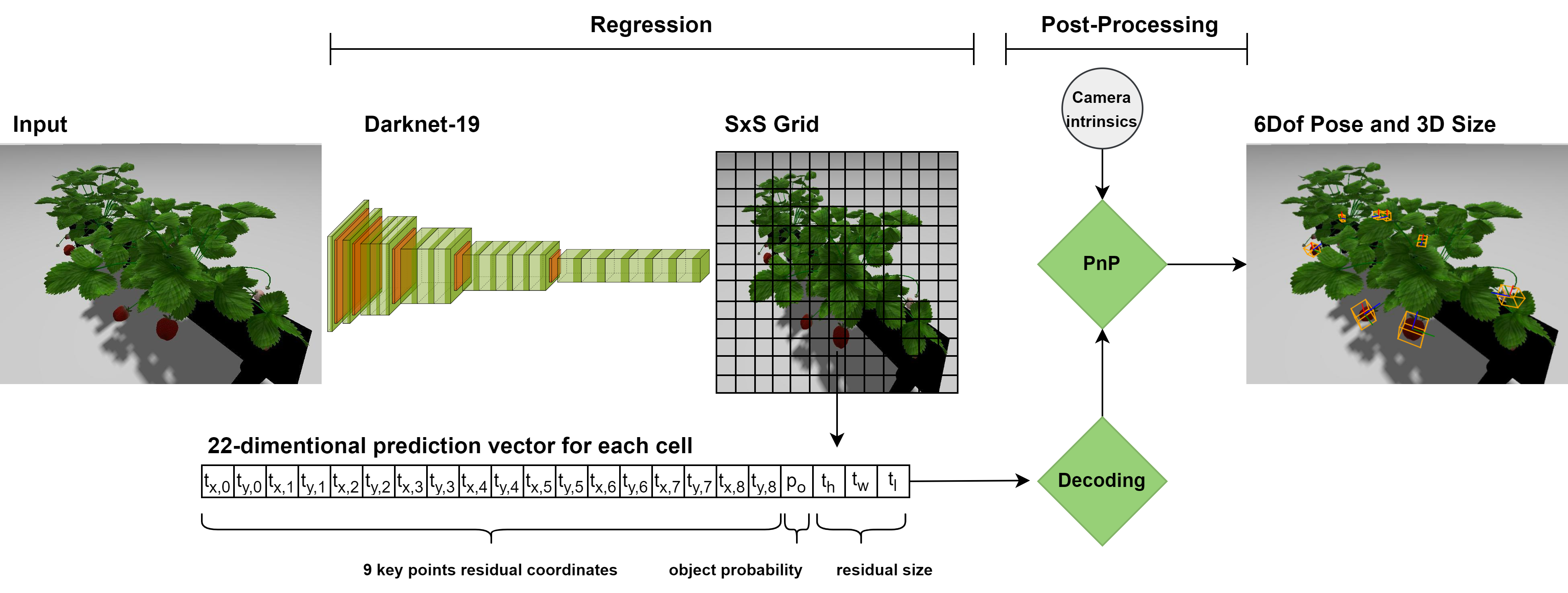}
\caption{Overview of our two-stage 6DoF pose and 3D size estimation method. Given an image, the neural network segments it into an SxS grid. Each grid predicts a 22-dimensional vector encompassing all information about the strawberry's 3D bounding box. Once decoded, the 3D size can be directly obtained, and the 6DoF pose can be computed using PnP algorithm, combined with the camera's intrinsic parameters.}
\label{fig2}
\end{figure*}

\subsection{Network Architecture}


Keypoints-based methods of 6DoF pose estimation are characterized by their speed, accuracy, and robustness. Concurrently, YOLO \cite{redmon2016you} remains a leading choice among 2D detectors for robotic fruit harvesting, as evidenced by recent research \cite{montoya2022vision}. Building upon these insights, and drawing inspiration from the foundational framework of YOLO, as well as its progressive successors, particularly YOLO3D \cite{ali2018yolo3d} and YOLO6D \cite{tekin2018real}, we have developed a novel two-stage method for keypoints-based 6DoF pose and 3D size estimation. This network is illustrated in Fig. \ref{fig2}.


The network accepts an RGB image of arbitrary resolution and processes it through the modified Darknet-19 network \cite{redmon2017yolo9000} to partition it into an $S\times S$ grid. Within each of these grid cells, the network predicts five anchors. Each of these anchors forecasts a 22-dimensional vector. This vector embodies the residual offset representation of one central point $[t_{x,0}\;t_{y,0}]^{\top}$, eight vertices $[t_{x,1\sim8}\;t_{y,1\sim8}]^{\top}$, and three dimensional sizes $[t_{w}\;t_{h}\;t_{l}]^{\top}$, as well as one strawberry confidence value $p_{o}$.


A strawberry, when represented in the 3D space, is encapsulated within a 3D bounding box. The 2D projection coordinates of the center point and eight vertices of this 3D bounding box can be determined using the following relations:
\begin{align}
x_{0} = \sigma(t_{x,0}) + c_{x} , \quad y_{0} = \sigma(t_{y,0}) + c_{y} \\
x_{1\sim8} = t_{x,1\sim8} + c_x , \quad y_{1\sim8} = t_{y,1\sim8} + c_y
\end{align}
Here, the function $\sigma()$ donates the sigmoid function that constrains the projected center point within its respective cell. Additionally, $c_x$ and $c_y$ represent the cell's coordinates.

For determining the dimensional parameters of the 3D bounding box, denoted by $[h\;w\;l]^\top$, the following formulae are employed:
\begin{align}
h = \overline{h}\cdot e^{t_h} , \quad w = \overline{w}\cdot e^{t_w} , \quad l = \overline{l}\cdot e^{t_l}
\end{align}
In this context, $e$ refers to the base of natural logarithms. The parameters $[\overline{h}\;\overline{w}\;\overline{l}]^\top$ are the average height, width and length of the strawberries that we pre-computed across the entire dataset. Considering the relatively consistent size variations observed in strawberries, this assumption is deemed appropriate.

Finally, leveraging the derived 3D bounding box details in tandem with the camera's intrinsic matrix $K$, we can recover the 6DoF pose of the strawberry relative to the camera's coordinate system with the PnP algorithm.

\subsection{Loss Function}

The loss function employed in our approach augments the foundational YOLO loss function and is segmented into three distinct components.

\noindent\textbf{Coordinate Loss:} This encapsulates the discrepancies in the 2D projected coordinates across nine points, and is computed using the mean square error:
\begin{equation}
L_{coord} = \sum_{i=0}^{S^2}\sum_{j=0}^{A}\sum_{k=0}^{9}\mathbf{1}_{ij}^{obj}[(x_k-\hat{x}_k)^2+(y_k-\hat{y}_k)^2] 
\end{equation}

\noindent\textbf{Dimension Loss:} The deviations in the 3D bounding box sizes are captured in this component, using the mean squared error: 
\begin{equation}
L_{dim} = \sum_{i=0}^{S^2}\sum_{j=0}^{A}\mathbf{1}_{ij}^{obj}[(h-\hat{h})^2+(w-\hat{w})^2+(l-\hat{l})^2] 
\end{equation}

\noindent\textbf{Confidence Loss:} Instead of using the 3D intersection over union (IoU) score, akin to the YOLO design, we adopt the idea mentioned in YOLO6D \cite{tekin2018real} to approximate the confidence score $C$ with the distance of the nine predicted key points from their true values.  
\begin{align}
L_{conf} = \sum_{i=0}^{S^2}\sum_{j=0}^{A}\mathbf{1}_{ij}^{obj}[(C-\hat{p_o})^2] \\
L_{conf\_no} = \sum_{i=0}^{S^2}\sum_{j=0}^{A}\mathbf{1}_{ij}^{noobj}[(C-\hat{p_o})^2]
\end{align}

Aggregating these individual losses, the overall loss is derived from a weighted combination:
\begin{equation}
\begin{split}
L_{all} = & \lambda_{coord}L_{coord} + \lambda_{dim}L_{dim} \\
          & + \lambda_{conf}L_{conf} + \lambda_{conf\_no}L_{conf\_no} \\
\end{split}
\end{equation}
Here $\lambda_{coord}=1$, $\lambda_{dim}=\lambda_{conf}=5$, and $\lambda_{conf\_no}=0.1$.

\noindent\textbf{Handling symmetry in pose estimation:}
Strawberries exhibit approximate symmetry along their main axis, stretching from the tip to the top. This characteristic implies that rotating the associated 3D bounding box around this symmetry axis results in similar projection views, yet generates different ground truth coordinates for these projections. 

To handle the inherent symmetry of strawberries, we have further incorporated the strategy from Wang et al. \cite{wang2019normalized}.  Every strawberry instance is rotated around its symmetry axis to generate a total of 12 ground truths. The final symmetric loss is thus computed as:
\begin{equation}
L_{all\_sym} = \min_{i=1\sim12}L_{all}(y^{(i)},\hat{y}) 
\end{equation}

\subsection{Synthetic Dataset Generation Pipeline}

\begin{figure*}
    \centering
    \begin{subfigure}{0.25\textwidth}
        \centerline{\includegraphics[width=0.6\linewidth]{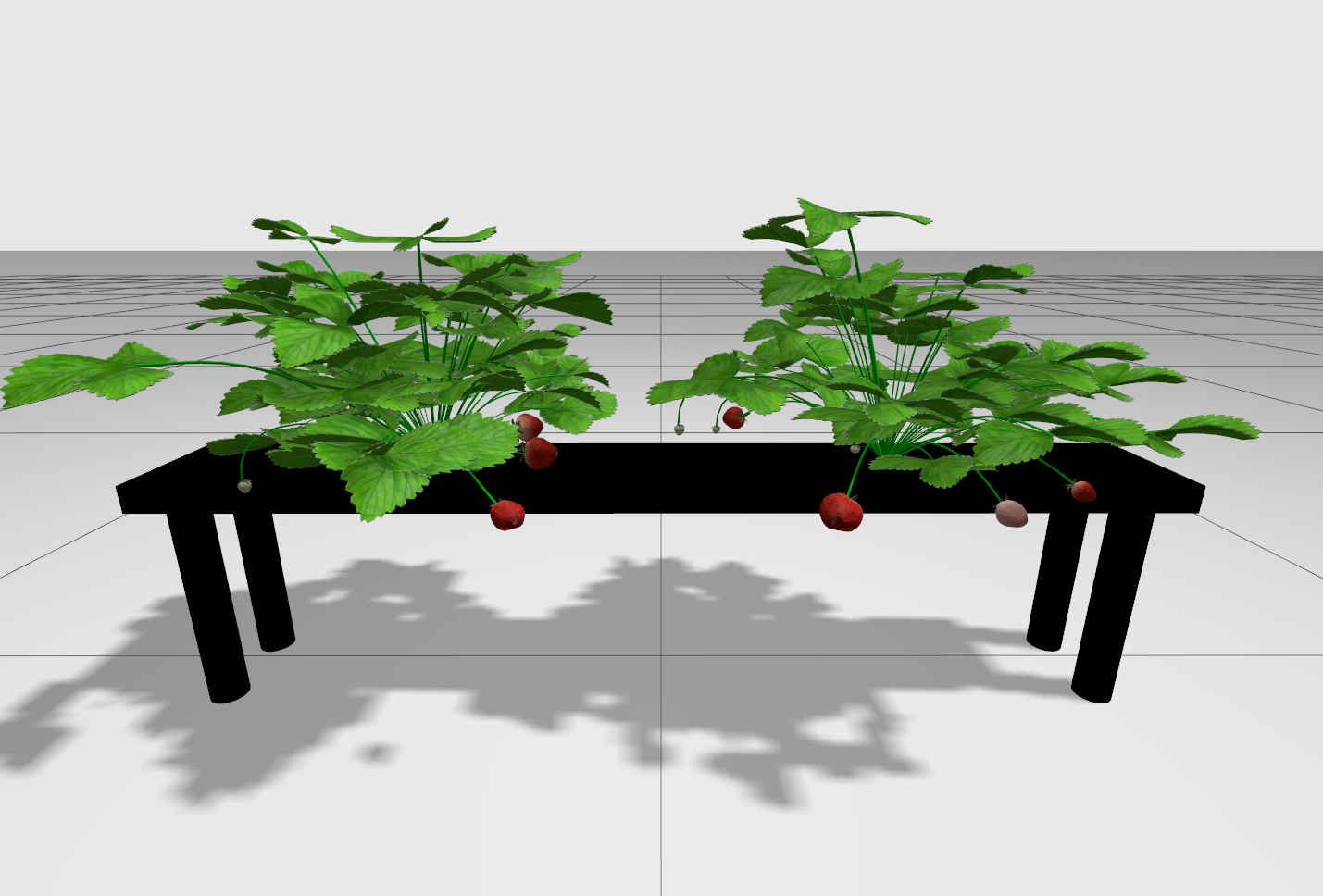}}
        \vspace{0.5pt}
        \centerline{\includegraphics[width=0.6\linewidth]{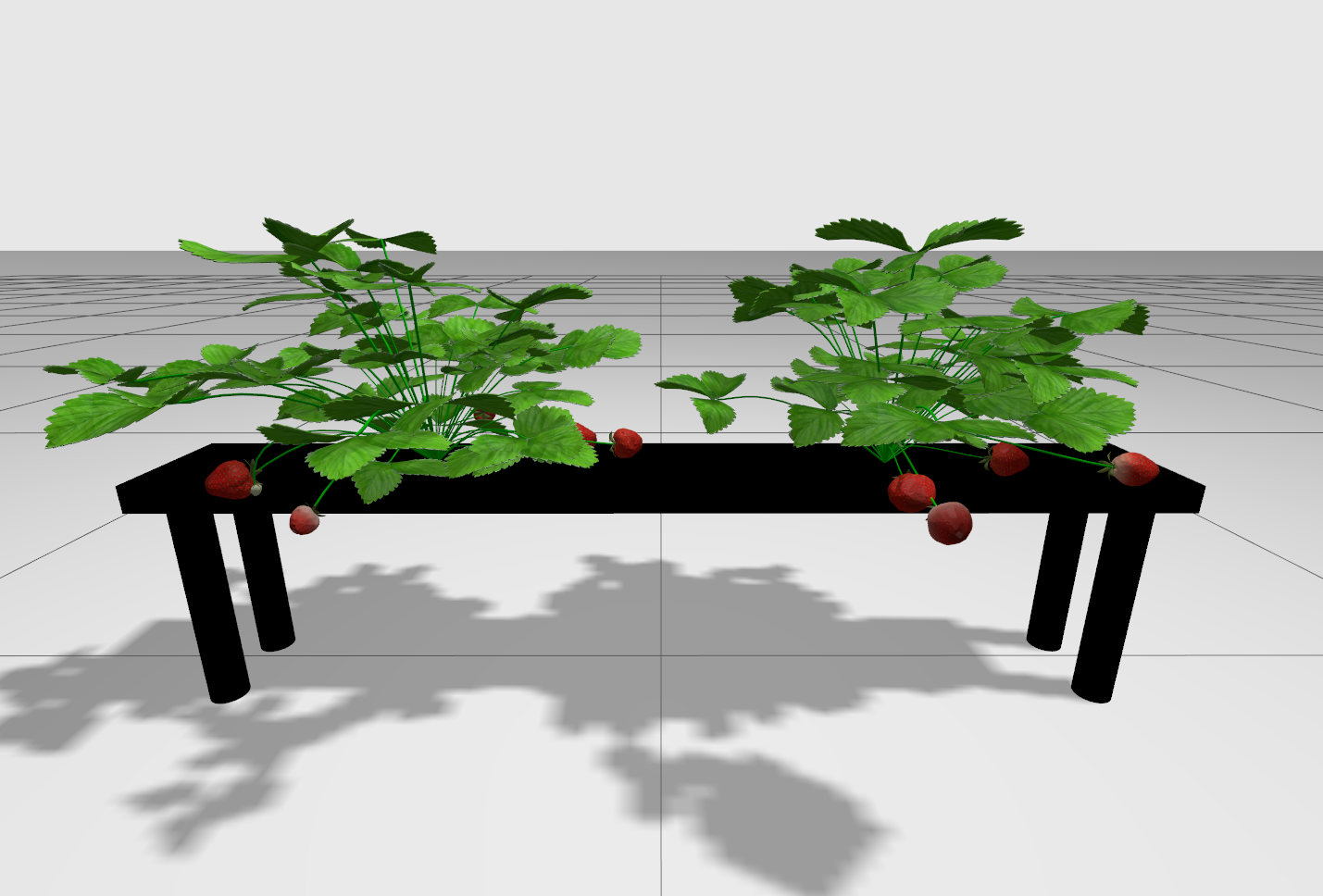}}
        \caption{}
        \label{fig:sub1a}
    \end{subfigure}%
    \begin{subfigure}{0.25\textwidth}
        \centerline{\includegraphics[width=0.6\linewidth]{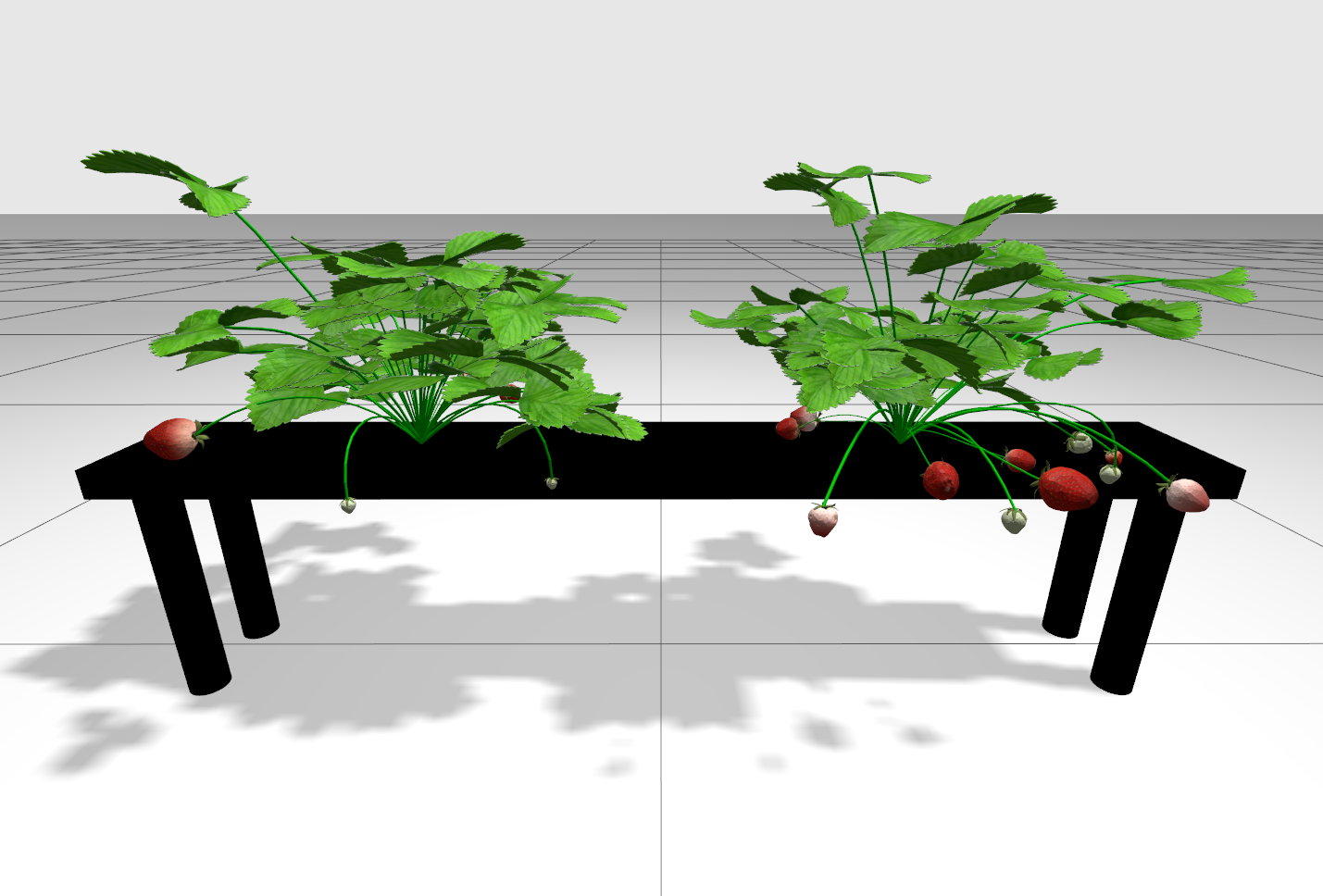}}
        \vspace{0.5pt}
        \centerline{\includegraphics[width=0.6\linewidth]{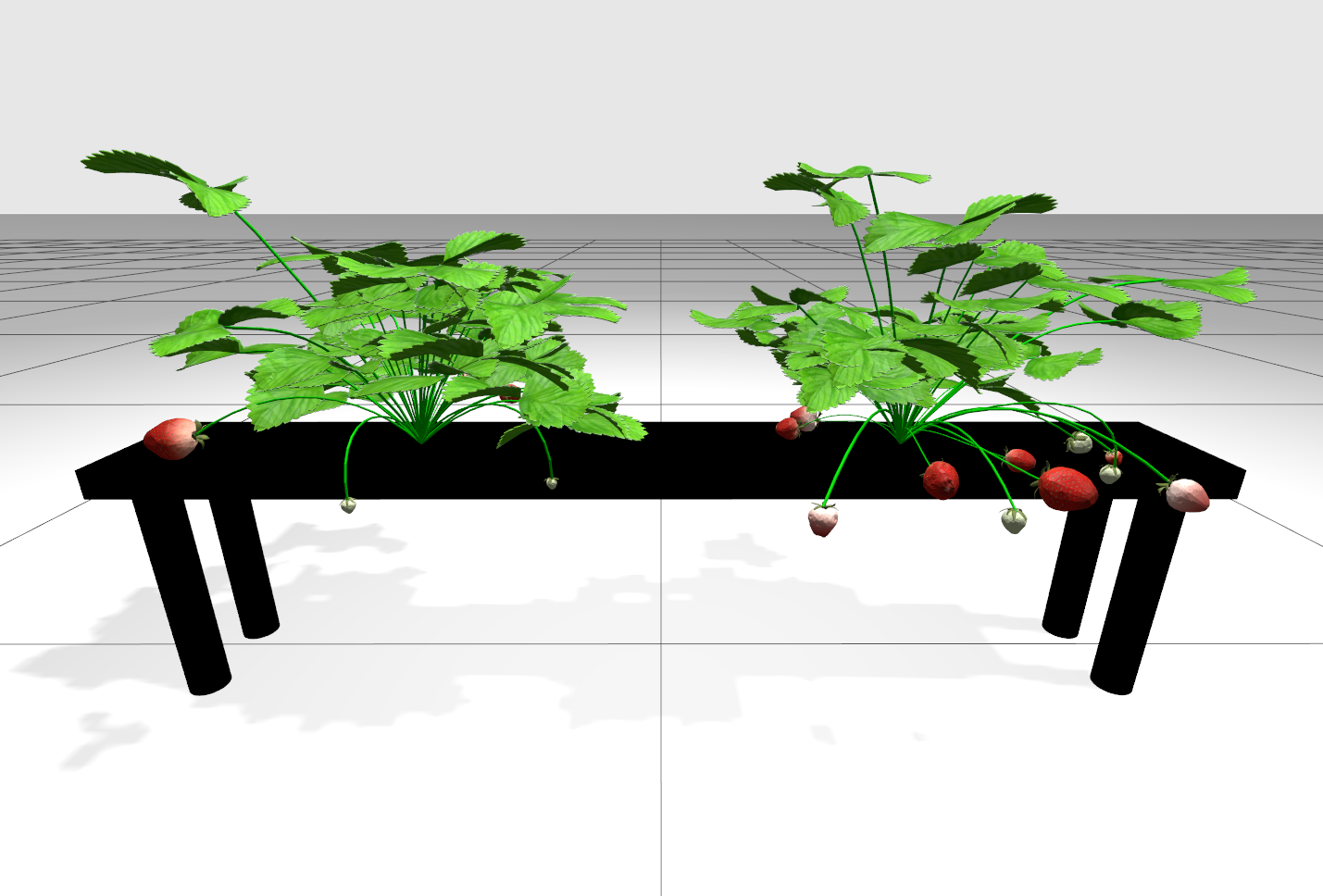}}
        \caption{}
        \label{fig:sub1b}
    \end{subfigure}%
    \begin{subfigure}{0.25\textwidth}
        \centerline{\includegraphics[width=0.6\linewidth]{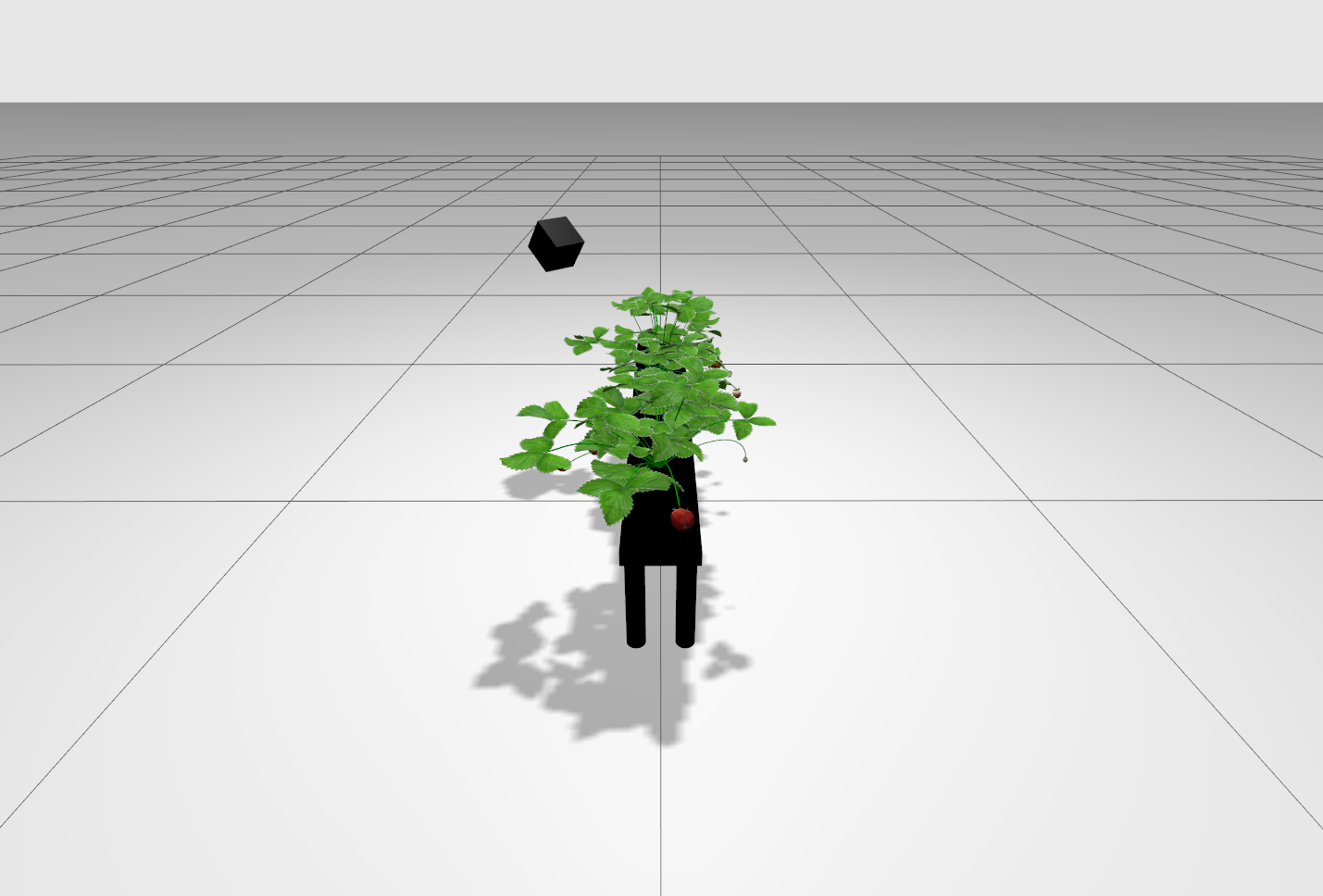}}
        \vspace{0.5pt}
        \centerline{\includegraphics[width=0.6\linewidth]{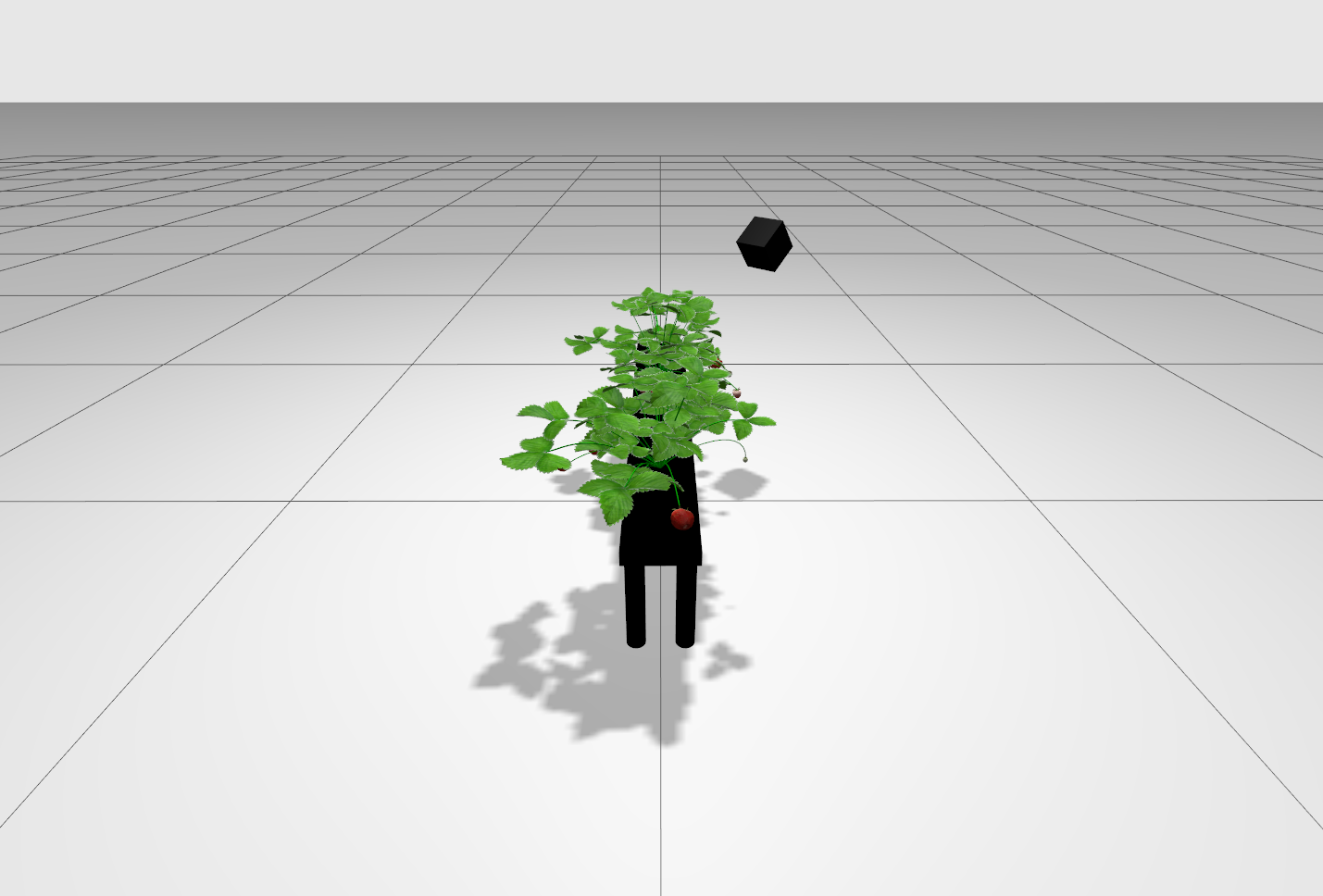}}
        \caption{}
        \label{fig:sub1c}
    \end{subfigure}%
    \begin{subfigure}{0.25\textwidth}
        \centerline{\includegraphics[width=0.6\linewidth]{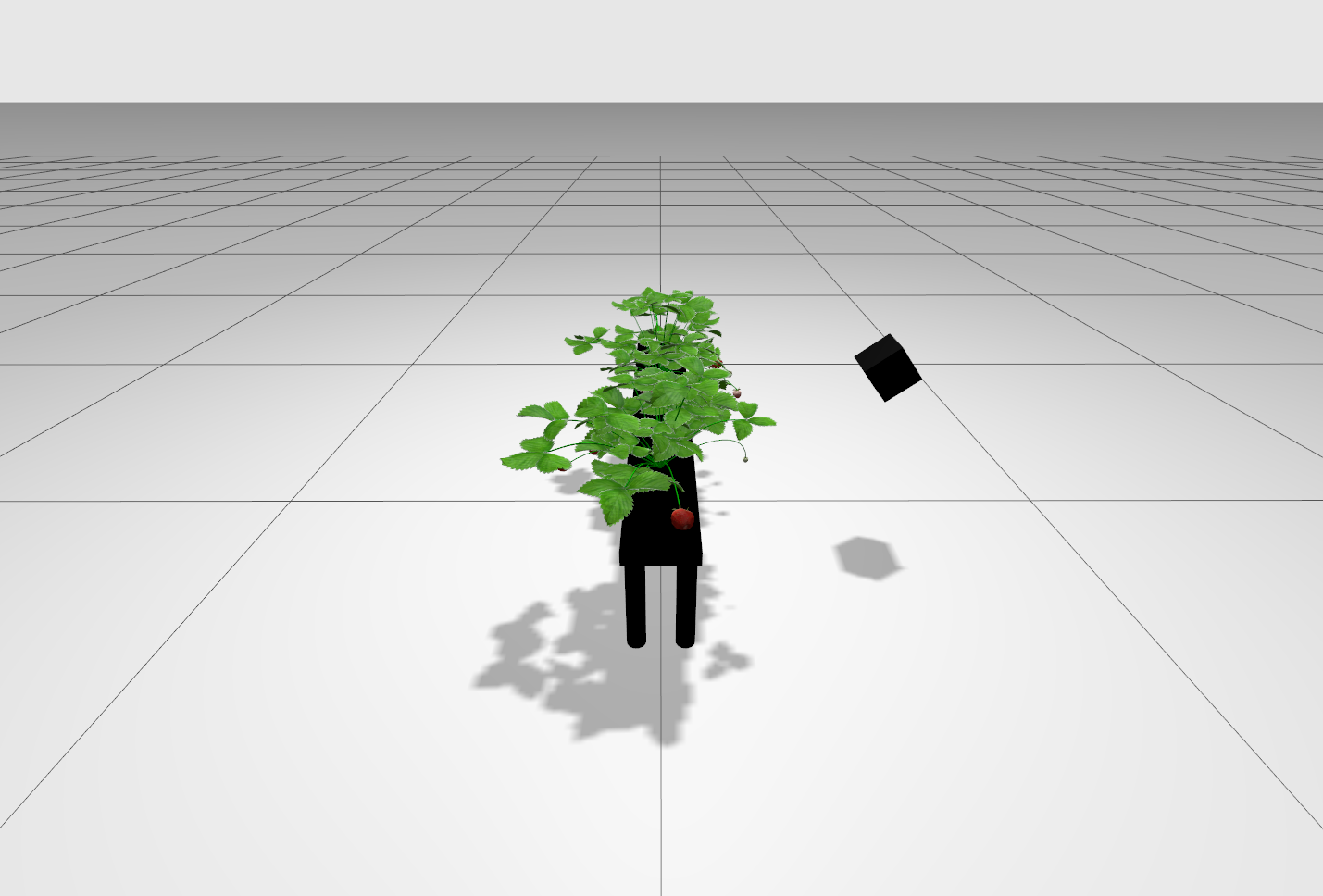}}
        \vspace{0.5pt}
        \centerline{\includegraphics[width=0.6\linewidth]{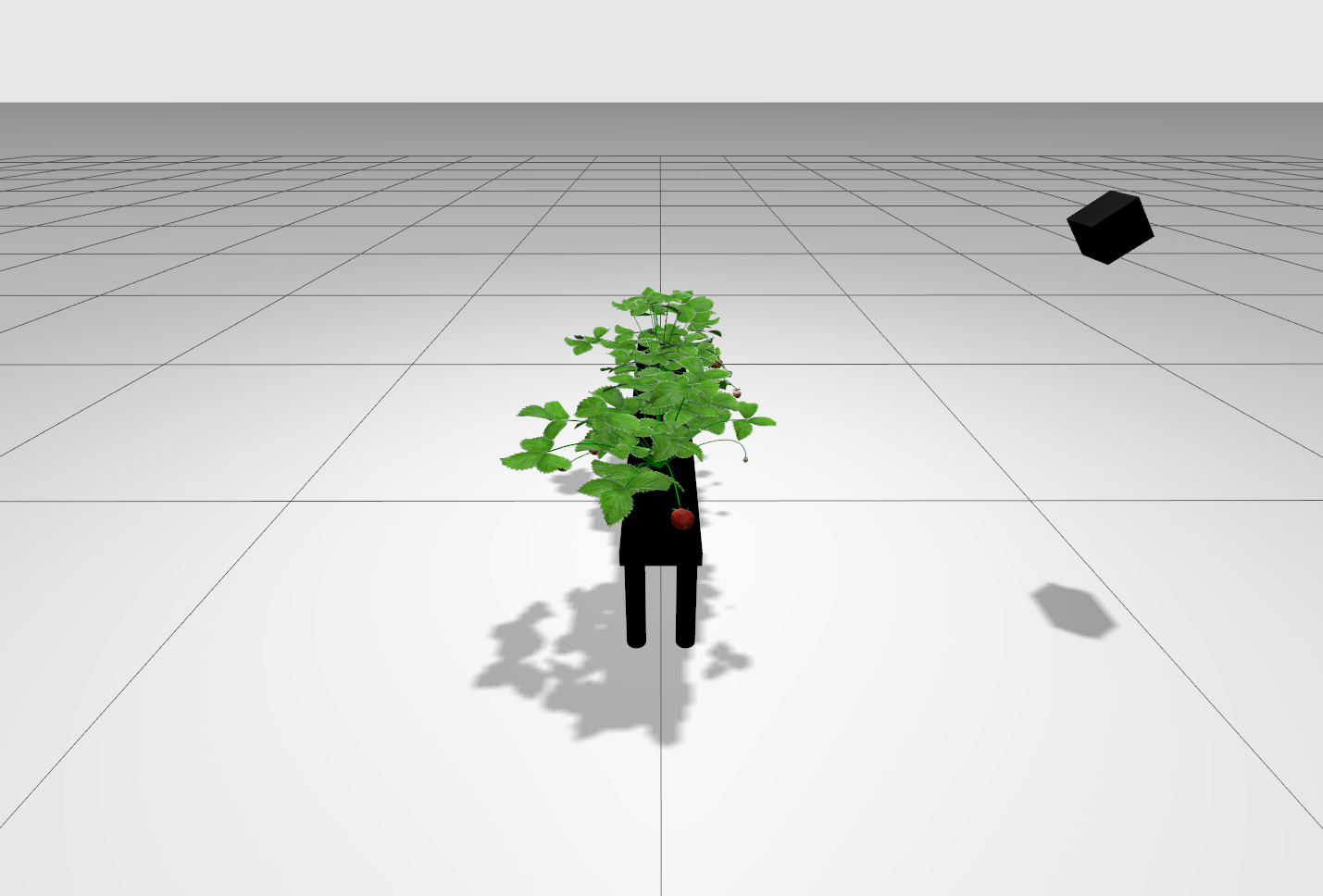}}
        \caption{}
        \label{fig:sub1d}
    \end{subfigure}
    
    \caption{Explanation of how we generate a diverse 6DoF pose and 3D size dataset for strawberries in the Ignition Gazebo simulator: (a) Each generated strawberry is random, encompassing variations in plant shape, the distribution of strawberries on the plant, and the size, ripeness, and pose of the strawberries. (b) For each batch of strawberries, different lighting conditions are set to mimic natural lighting variations. (c) The camera randomly selects angles from a reasonable range to capture images. (d) For a given camera angle, the distance between the camera and the strawberries is also adjusted. The entire process is automated and continues until the preset sample size is reached. The final dataset, named Straw6D, includes RGB images, depth images, 3D bounding box annotations, strawberry instance segmentation masks and point clouds.}
    \label{fig3}
\end{figure*}

For the training of our model, a large-scale strawberry dataset labeled with 6DoF pose and 3D size parameters is essential. Given the inherent lack of precise natural perspectives, manually annotating 6DoF poses and 3D sizes of target objects in 2D images is a notably challenging endeavor. To the best of our knowledge, there is no existing strawberry dataset that offers public access to both 6DoF and 3D size estimations. However, with advancements in computer graphics and simulators, we opted to generate a synthetic categorical 6DoF Pose dataset for strawberries from a simulated environment, named Straw6D. This dataset encompasses all the annotation information necessary for both training and testing our strawberry pose estimator.

We developed our synthetic dataset generation pipeline in the Ignition Gazebo simulator to integrate our other modules within a ROS-centric framework. 
Ignition Gazebo simulator is the successor to classic Gazebo simulator, which has improved visual fidelity to enhance its suitability for computer vision systems. 
We utilized the 3D bounding box camera feature that can directly generate precise three-dimensional state information of specified objects from rendered images of the simulation camera, even when the strawberries are partially obscured. This capability allows us to automatically annotate the image stream from the camera. 
To provide a comprehensive dataset suitable for a wide range of research applications, we enriched our dataset with other 
features of Ignition Gazebo, incorporating depth images, strawberry instance segmentation masks, and point clouds.

We adopted the SDF strawberry model and its instance randomization script created by Sather et al. \cite{sather2019viewpoint} for our simulations.
Diverging from merely simulating a single strawberry plant, we crafted additional scripts
to automate the insertion of multiple strawberry plants and fruit instances into the Ignition Gazebo simulator. 
As a result, each insertion involves prior randomization, affecting  the number of fruits, their size scaling, pose, and distribution on the plant, alongside the variability in the plant's root, stem, and leaf configurations. 

To mitigate the common issue of models trained on synthetic datasets underperforming with real-world images, we implemented domain randomization techniques \cite{tobin2017domain}, strategically varying camera settings and introducing random lighting conditions within the simulator. This approach aimed to mimic the variability of real-world conditions, thereby enhancing the dataset's diversity with cluttered scenes, varied plant forms, and randomized berry positions and sizes, ensuring the test scenarios distinctly differ from the training set. This methodology, illustrated in Fig. \ref{fig3}, is not restricted to strawberries alone but is adaptable for generating datasets for other objects, such as apples, necessitating 6DoF pose estimation.

For each sampled data instance of the strawberry, there is a 3D bounding box-related annotation. This annotation comprises three-dimensional translation, three-dimensional rotation, and three-dimensional size for length, width, and height, as well as the strawberry maturity.

\section{Experimental Results}


\subsection{Experimental Setups}

To train our strawberry pose estimator, we auto-generated a total of 2000 labeled images from the Gazebo Ignition simulator. Out of the entire dataset, 80\% images were used for training and the remaining 20\% were set aside for testing.

We trained the network on an NVIDIA GeForce RTX 3060Ti GPU using a batch size of 8 over 600 epochs, beginning with pretrained weights sourced from ImageNet\cite{deng2009imagenet}. Our data augmentation techniques encompassed random flipping, scaling, cropping, and color jittering. For optimization, we selected the SGD algorithm, setting an initial learning rate at 1e-4. This rate was then reduced tenfold at the 120th and 240th epochs. Our developmental platform comprised Python 3.8, Pytorch 2.2.0, and CUDA 12.1.

\subsection{Performance on Simulated Data}

In this study, we utilized two widely adopted metrics to quantitatively assess the performance of our proposed model on the synthetic dataset. 
For the evaluation of 3D detection and object dimension estimation, we used 3D Intersection over Union (IoU) \cite{hou2020mobilepose}, which calculates the overlap of the 3D bounding boxes corresponding to the predicted and true poses. A prediction is considered correct if the overlap ratio exceeds a specified threshold. Since the precision results can vary depending on the choice of the threshold, we selected multiple thresholds of 0.5, 0.6, 0.7, and 0.8 for measurement.  
For the evaluation of 6DoF pose estimation, we used rotational and translational errors, directly computing the differences in rotation and translation between the predicted and true poses. A prediction is deemed correct if the translational error is below a distance threshold and the rotational error is below an angular threshold at the same time. We measured these with threshold sets at ($1$cm, $10^{\circ}$), ($1$cm, $20^{\circ}$), ($2$cm, $10^{\circ}$), and ($2$cm, $20^{\circ}$).

\begin{figure}[!t]
\centerline{
\includegraphics[width=0.24\textwidth]{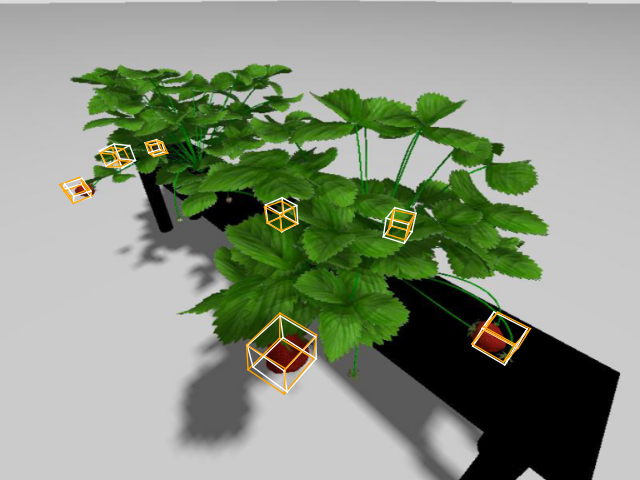}
\includegraphics[width=0.24\textwidth]{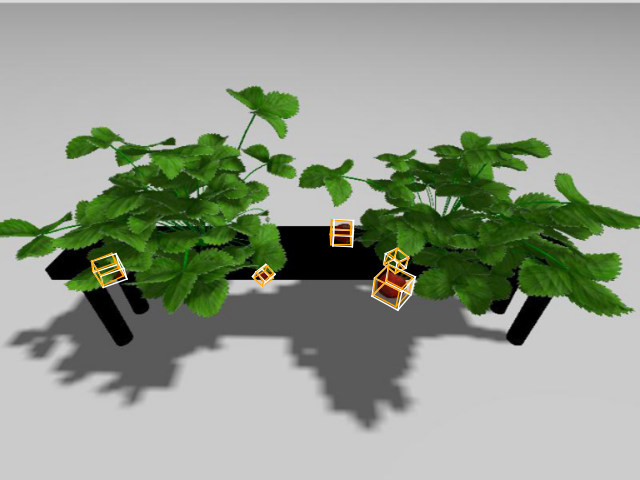}
}
\caption{Inferences made on the synthetic test dataset. The orange 3D bounding boxes represent our predictions of the strawberries, as well as the white ones denote the ground truths.}
\label{fig4}
\end{figure}

Through the model’s prediction results combined with the PnP algorithm, we obtained the estimated rotation and translation matrices. In a similar vein, we derived the transformation matrix for the ground truth. Subsequently, rotational and translational errors are acquired from the discrepancies of these two transformation matrices. Thereafter, we back-projected both the estimated and ground truth 2D projected coordinates of the center point and vertices to 3D space, calculating the 3D IoU value. Further, due to the symmetrical assumption of the strawberry, we selected the highest score as well as the lowest errors among the 12 ground truths to avoid penalization. Table \ref{tab1} summarizes the results of 3D detection and object dimension estimation. Table \ref{tab2} summarizes the results of 6DoF pose estimation. The inference outcomes on the test dataset can be observed in Fig.~\ref{fig4}. In terms of inference speed, when tested on the same GPU as training, on average we achieved a runtime of 16.60 ms, leading to 60 FPS. By increasing the level of thresholds, we observed a significant performance drop.
This indicates room for enhancement in our model's precision. Opting for a larger model architecture might be beneficial, even if it results in sacrificing some computational speed.

\begin{table}[htbp]
\caption{Model Performance in terms of AP of 3D IoU for strawberries detection and dimension estimation. The metric unit is \%.}
\centering
\begin{tabular}{l|rrrrr}
\toprule
\textbf{Threshold} & 0.5 & 0.6 & 0.7 & 0.8 \\
\midrule
\textbf{Average Precision (\%)} & 84.77 & 78.6 & 45.78 & 19.87 \\
\bottomrule
\end{tabular}
\label{tab1}
\end{table}

\begin{table}[htbp]
 \caption{Model Performance in terms of translational and rotational errors for 6DoF pose estimation. The metric unit is \%.}
\centering
\resizebox{\linewidth}{!}{
\begin{tabular}{l|rrrrr}
\toprule
\textbf{Threshold} & (2cm, 20°) & (2cm, 10°) & (1cm, 20°) & (1cm, 10°) \\
\midrule
\textbf{Average Precision (\%)} & 74.73 & 44.09 & 61.95 & 35.58 \\
\bottomrule
\end{tabular}
}
\label{tab2}
\end{table}

\subsection{Sim-to-real Evaluation on Real Data}

While utilizing synthetic data for training presents a potential solution to one of the primary bottlenecks in current deep learning models—namely, the need for large-scale annotated datasets—it inevitably faces the challenge of sim-to-real transfer. 
For the strawberry harvesting task, we aim to leverage available real RGB image datasets to make our approach suitable for real-world scenarios. We first use real strawberries datasets to learn deep feature representations, and then fine-tune the model to acquire 3D knowledge using the proposed synthetically generated 6DoF pose strawberries dataset.

\begin{figure}[htpb]
\centerline{
\includegraphics[width=0.24\textwidth]{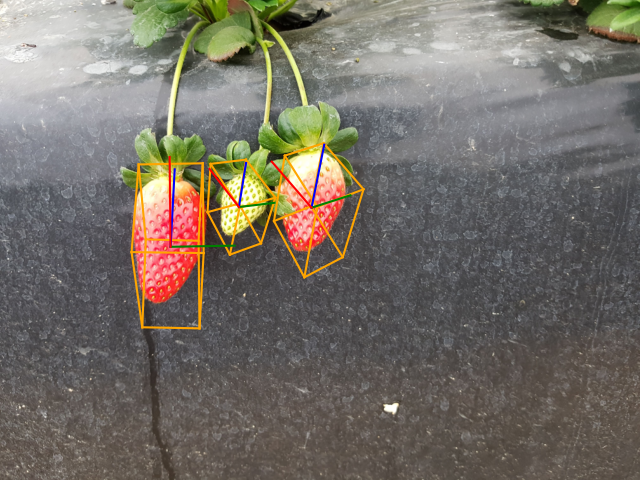}
\includegraphics[width=0.24\textwidth]{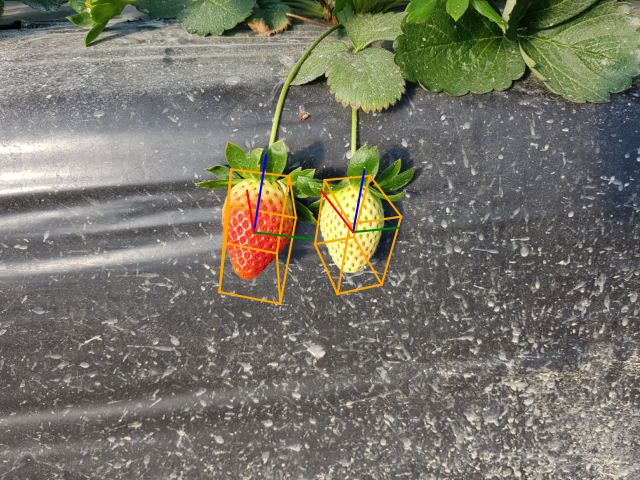}
}
\vspace{3pt}
\centerline{
\includegraphics[width=0.24\textwidth]{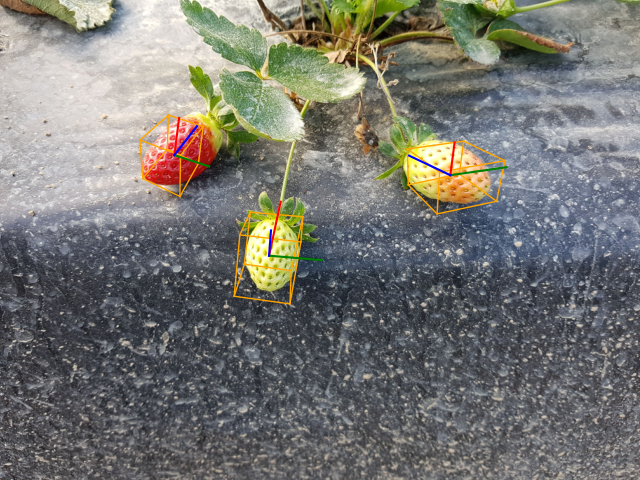}
\includegraphics[width=0.24\textwidth]{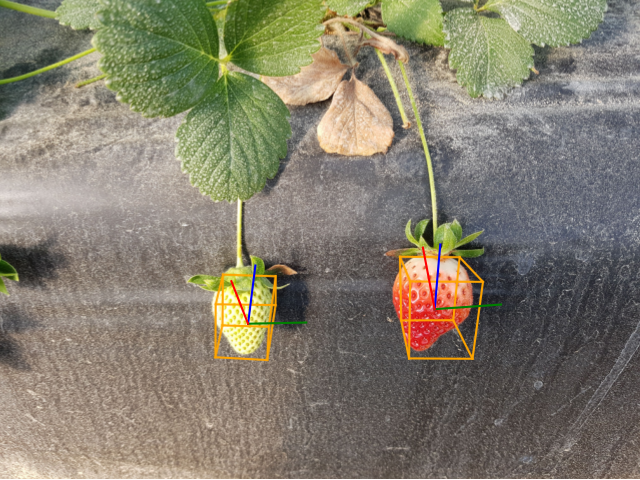}
}

\caption{Inferences made on the real-world strawberry dataset~\cite{PEREZBORRERO2020105736}. The orange 3D bounding boxes represent our predictions of the strawberries}
\label{fig5}
\end{figure}

Towards this goal, we employed the StrawDI dataset \cite{PEREZBORRERO2020105736} to train our model and test its sim-to-real performance. 
We pretrained the backbone of our model on the StrawDI dataset by adding a YOLO 2D detection head. Subsequently, we further trained the complete network on Straw6D. During the training process, we initially trained the whole network for 120 epochs and then froze all parameters of the backbone, and then fine-tuned the remaining parts of the model for an additional 480 epochs. 
The resultant model demonstrated accuracy comparable to that of models trained solely in simulated environments when evaluated on a synthetic data test set. However, its performance on real-world datasets significantly surpassed that of models trained exclusively in simulation.
Furthermore, the results we obtained are interesting, particularly because the synthetic dataset lacks large unripe strawberries, whereas the real dataset contains many instances of them. Remarkably, our model demonstrated the capability to detect strawberries irrespective of their ripeness level.
The inference results are depicted in Fig.~\ref{fig5}.


\section{Conclusion}


In this paper, we have studied the potential of a lightweight 6D pose and size estimator for strawberry harvesting task. Our approach was built based on the YOLO framework, which we adapted to become a 6D pose estimator by infusing additional regression terms,
accounting for the fruit's inherent symmetry. To further enrich our approach, a comprehensive 6D strawberries pose and size dataset was generated using Gazebo simulator. This dataset considered factors like random lighting, strawberry position and size, and variations in camera distances and perspectives for sim2real gap. 
Experimental results showed that the proposed approach worked well in both simulation and real RGB data. 
In our future research work, 
we plan to develop a dual-arm strawberry picking system to address the challenge of harvesting ripe strawberries hidden behind unripe ones. 




\bibliographystyle{IEEEtran}
\bibliography{reference}

\end{document}